\documentclass[runningheads]{llncs}
\usepackage{graphicx}
\usepackage{comment}
\usepackage{amsmath,amssymb} % define this before the line numbering.
\usepackage{color}
\usepackage{siunitx}
\usepackage{mathtools}
\usepackage{multirow}
\usepackage{booktabs}
\usepackage{subcaption}
\usepackage{gensymb}
\usepackage{colortbl}
\usepackage{arydshln}
\usepackage{multicol}
\usepackage{ifpdf}

\usepackage{hyperref}
\hypersetup{
    colorlinks=true,
    linkcolor=blue,
    filecolor=cyan,      
    urlcolor=magenta,
}

\usepackage[linesnumbered,ruled,vlined]{algorithm2e}

\begin{document}
% \linenumbers
\pagestyle{headings}
\mainmatter
\def\ECCVSubNumber{206}  % Insert your submission number here

\title{OnlineAugment: Online Data Augmentation with Less Domain Knowledge} % Replace with your title

%\begin{comment}
% \titlerunning{ECCV-20 submission ID \ECCVSubNumber} 
% \authorrunning{ECCV-20 submission ID \ECCVSubNumber} 
% \author{Anonymous ECCV submission}
% \institute{Paper ID \ECCVSubNumber}
%\end{comment}
%******************

% \begin{comment}
\titlerunning{OnlineAugment}
\author{Zhiqiang Tang$^1$\thanks{Corresponding author: Zhiqiang Tang}\and Yunhe Gao$^1$\and Leonid Karlinsky$^2$\and Prasanna Sattigeri$^2$\and Rogerio Feris$^2$\and Dimitris Metaxas$^1$}
\authorrunning{Tang et al.}
\institute{Rutgers University, 
\email{\{zhiqiang.tang, yunhe.gao, dnm\}@rutgers.edu}\\
\and
IBM Research AI,
\email{\{LEONIDKA@il, psattig@us, rsferis@us\}.ibm.com}}
% \end{comment}
%******************
\maketitle

\begin{abstract}

Data augmentation is one of the most important tools in training modern deep neural networks. Recently, great advances have been made in searching for optimal augmentation policies in the image classification domain. However, two key points related to data augmentation remain uncovered by the current methods. First is that most if not all modern augmentation search methods are \textit{offline} and learning policies are isolated from their usage. The learned policies are mostly constant throughout the training process and are \textit{not adapted} to the current training model state. Second, the policies rely on class-preserving image processing functions. Hence applying current offline methods to new tasks may require domain knowledge to specify such kind of operations. In this work, we offer an orthogonal \textit{online} data augmentation scheme together with three new augmentation networks, co-trained with the target learning task. It is both more efficient, in the sense that it does not require expensive offline training when entering a new domain, and more adaptive as it adapts to the learner state. Our augmentation networks require less domain knowledge and are easily applicable to new tasks. Extensive experiments demonstrate that the proposed scheme alone performs on par with the state-of-the-art offline data augmentation methods, as well as improving upon the state-of-the-art in combination with those methods. Code is available at \url{https://github.com/zhiqiangdon/online-augment}. 

% Data augmentation is one of the most important tools in training of modern deep learning approaches. It is essential in the situation where only limited data is available for the target task, but is also crucial for generalization performance in case the data is abundant. Great advances have been recently made in the field of searching for optimal augmentation policies that brought significant improvements to state-of-the-art performance on various tasks. However, several key points related to data augmentation remain uncovered by the current methods. First is that most if not all modern augmentation search methods are \textit{offline} - relying on a large amount of GPU processing time. Hence it is difficult to adapt them to any particular new task. Second, the augmentation policies are mostly constant throughout the training process and are \textit{not adapted} to the current training model state. In this work, we offer the first \textit{online} method for augmentation search based on a set meta-trained augmentation networks, co-training with the target learning task. It is both more efficient, in the sense that it does not require expensive offline training, and more adaptive - as it adapts to the learning task at hand as well as to the state of the learner. Extensive experiments demonstrate that the proposed scheme on its own performs on par with the state-of-the-art augmentation search methods, as well as improving upon the state-of-the-art in combination with those methods.

% \keywords{Data Augmentation, Domain Knowledge, Adversarial Training.}
\end{abstract}
% \vspace{0.01pt}
\section{Introduction}
Data augmentation is widely used in training deep neural networks. It is an essential ingredient of many state-of-the-art deep learning systems on image classification \cite{simard2003best,krizhevsky2012imagenet,devries2017dataset,zhang2017mixup}, object detection \cite{girshick2018detectron,dai2016r}, segmentation \cite{fu2019dual,ronneberger2015u,yang2018denseaspp}, as well as text classification \cite{yang2019xlnet}. Current deep neural networks may have billions of parameters, tending to overfit the limited training data. Data augmentation aims to increase both the quantity and diversity of training data, thus alleviates overfitting and improves generalization.

Traditionally, data augmentation relies on hand-crafted policies. Designing the polices is usually inspired by domain knowledge and further verified by testing performance \cite{simonyan2014very,he2016deep}. For example, the typical routine in training CIFAR classifiers uses random cropping and horizontal flip to conduct data augmentation. Intuitively, these operations do not change the image labels, and they can also improve testing performance in practice. 

Recently, AutoML techniques \cite{zoph2016neural,baker2016designing} are used to automate the process of discovering augmentation polices. The resulted approaches, such as AutoAugment \cite{cubuk2019autoaugment} and its variants \cite{krizhevsky2012imagenet,zoph2019learning,lim2019fast,ho2019population} are quite successful and achieve state-of-the-art results.  We name them  {\it offline data augmentation} since the policy learning and usage are isolated. Moreover, these approaches use pre-specified image processing functions as augmentation operations. Defining the basic operations requires domain knowledge, which may impede their applications to more tasks.

% reinforcement based learning approach to discover domain specific optimal policies for data augmentation
%Recently, AutoAugment, using AutoML, can automate the process of discovering effective augmentation policies. AutoAugment actively explores policies and validates their effectiveness by repeatedly training models from scratch. Due to the large search space, the searching process, based on reinforcement learning, severely suffers from high computational demand. Subsequent works have speeded up its learning efficiency using different searching strategies. However, they all follow the formulation of AutoAugment that first searches policies using a sampled small dataset, and then applies them to the final large dataset. We name them as offline data augmentation since the policy learning and usage are isolated. Moreover, they all use pre-specified image processing functions as augmentation operations. Defining the basic operations requires domain knowledge, which may impede their applications to more tasks.

In this paper, we propose {\it OnlineAugment}, which jointly optimizes data augmentation and target network training in an online manner. The merits of OnlineAugment lie in three-fold. First, it is  orthogonal to the offline methods. Their complementary nature makes it possible to apply them together. Second, through the online learning, the augmentation network can adapt to the target network through training from the start to the end, saving it from the inconveniences of pre-training \cite{lee2019learning} or early stopping \cite{ratner2017learning}. Third, it is easy to implement and train OnlineAugment. In contrast, learning offline policies usually rely on distributed training, as there are many parallel optimization processes.

% Compared with the offline methods, ours is more efficient and easy-to-use. Learning offline policies usually relies on distributed training, as there are hundreds or thousands of parallel optimization processes. Fortunately, our online augmentation, with a single optimization procedure, requires only regular implementation and training. Besides, OnlineAugment is easy-to-use. Learning offline policies usually relies on distributed training, as there are hundreds or thousands of parallel optimization processes. Fortunately, our online augmentation, with a single optimization procedure, requires only regular implementation and training.

Furthermore, we propose more general data augmentation operations with less domain knowledge. Instead of using pre-defined image processing functions, such as rotation and color, we design neural networks to perform data augmentation. Specifically, we devise three learnable models: augmentation STN (A-STN), deformation VAE (D-VAE), and Perturbation VAE (P-VAE). It is nontrivial to craft STN \cite{jaderberg2015spatial} and VAE \cite{kingma2013auto} to conduct data augmentation. We also propose new losses to regularize them in training. Besides, OnlineAugment integrates both adversarial training and meta-learning in updating the augmentation networks. Adversarial training is to prevent overfitting, whereas meta-learning encourages generalization.

% First, we categorize the operations into three types: global spatial transformation, local deformation, and texture perturbation. Then we devise new augmentation networks inspired by the spatial transformer network (STN) \cite{jaderberg2015spatial} and variational auto-encoder (VAE) \cite{kingma2013auto}. For effective and efficient online training, our method embodies the spirits of adversarial training and meta-learning. Adversarial training is to prevent overfitting to training data, whereas meta-learning encourages generalization to validation data.

In summary, our key contributions are:

% \begin{itemize}
%     \item We propose a new online data augmentation scheme, which is orthogonal to the previous offline ones such as state-of-the-art AutoAugment. Experiments on CIFAR, SVHN, and ImageNet show that OnlineAugment achieves comparable performances to AutoAugment. More excitingly,  OnlineAugment can further boost state-of-the-art performances if used together with AutoAgument policies.
%     \item To reduce the dependency on domain knowledge, we devise more data augmentation models rather than using image processing functions. Three models are designed to generate augmentations complementary to each other. To integrate different types of augmentations, we inject multiple groups of batch normalization layers into the target network, each of which is in charge of one type of augmentation.
%     \item Our online data augmentation is efficient without any separate policy searching. We take advantage of adversarial training and meta-learning to conduct efficient online training. Table ? shows the comparisons of search space and time costs with offline augmentation methods.
%     \item The proposed OnlineAugment is general for different tasks. In addition to the experiments on classification datasets CIFAR, SVHN, and ImageNet, we also validate OnlineAugment in a new liver segmentation task. OnlineAugment significantly advances the IOU by \%.
% \end{itemize}{}

% Leonid's variant
\begin{itemize}
    \item We propose a new online data augmentation scheme based on meta-learned augmentation networks co-trained with the target task. Our framework is complementary to the state-of-the-art offline methods such as AutoAugment. Experiments on CIFAR, SVHN, and ImageNet show that on its own, OnlineAugment achieves comparable performances to AutoAugment. More excitingly, OnlineAugment can further boost state-of-the-art performances if used jointly with AutoAgument policies.
    \item We propose three complementary augmentation models responsible for different types of augmentations. They replace the image processing functions commonly used in contemporary approaches and make our method both more adaptive and less dependent on domain knowledge.
    % Our framework is both more adaptive to the target task and to the state of the learner, as well as it 
    % \item Our online data augmentation is efficient without any separate policy searching. We take advantage of adversarial training and meta-learning to conduct efficient online training. Table ? shows the comparisons of search space and time costs with offline augmentation methods.
    \item We show that the proposed OnlineAugment can generalize to tasks different from object classification by applying it to a liver\&tumor segmentation task, demonstrating improved performance compared with the state-of-the-art RandAugment on this task.
\end{itemize}{}

\section{Related Work}

Data augmentation has been shown to improve the generalization of machine learning models and is especially effective in training deep neural networks. It is essential in the situation where only limited data is available for the target task, but is also crucial for generalization performance in case the data is abundant.

% overparametrized
% Most of the popular approaches generate class preserving perturbations of the existing, often limited, training samples. 

% More general approaches aim to learn full data distribution from limited samples, which can then be used to generate unlimited training samples.

Known class-preserving transformation has been routinely used to expand labeled datasets for training image classifiers. These include operations such as cropping, horizontal and vertical flips, and rotation \cite{ciregan2012multi,simard2003best,krizhevsky2012imagenet}. Recently, reinforcement learning has been used to learn the optimal sequence of such transformations for a given dataset that leads to improved generalization \cite{ratner2017learning}. AutoAugment \cite{cubuk2019autoaugment} falls under this category of methods and actively explores policies and validates their effectiveness by repeatedly training models from scratch. Due to the large search space, the searching process, based on reinforcement learning, severely suffers from high computational demand. Subsequent works \cite{cubuk2019randaugment,ho2019population} in this direction have been aimed at reducing the computational complexity of the search strategies. However, they all follow the formulation of AutoAugment that first searches policies using a sampled small dataset, and then applies them to the final large dataset. Thus the policy learning and usage are isolated. Adversarial AutoAugment \cite{zhang2019adversarial} jontly optimizes the polices and target learner. However, the learned policies are still based on domain-specific image processing functions. 

More general transformations, such as Gaussian noise and dropout, are also effective in expanding the training set \cite{srivastava2014dropout,taylor2017improving,devries2017improved}. Spatial Transformer Network (STN) can perform more adaptive transformations than image processing functions such as rotation. However, it was designed for localization, not for data augmentation. In this paper, we craft it to conduct data augmentation. Generative models are also helpful for data augmentation. DAGAN \cite{antoniou2017data} employs a Generative Adversarial Network (GAN) \cite{goodfellow2014generative} to learn data augmentation for few-shot classification. In this work, we devise two augmentation models based on Variational Auto-encoder (VAE) \cite{kingma2013auto}, as another popular generative model.

% Transformations that are less domain-specific less such adding Gaussian noise, dropout, or patch-level occlusion are also effective in expanding the training set \cite{srivastava2014dropout,taylor2017improving,devries2017improved}. Strategies such as \textit{mixup} \cite{zhang2017mixup} alternatively use simple combinations of the training samples to create labeled augmentations. In \cite{antoniou2017data}, the authors employ a Generative Adversarial Network (GAN) to learn adaptive class preserving transformations, which are then used during a disjoint classifier training process.

Adversarial training \cite{goodfellow2014explaining,kurakin2016adversarial,madry2017towards} can serve as a general data augmentation framework. It aims to generate adversarial perturbations on input data. The adversarial examples are further used in training models to improve their robustness. It has been shown that adversarial training can hurt model generalization although it can boost robustness \cite{madry2017towards,xie2019feature}. Concurrent work AdvProp \cite{xie2019adversarial} successfully adapts adversarial training to advance model generalization. It uses the common adversarial attack methods, such as PGD and I-FGSM, to generate additive noises. In contrast, our models can learn to generate more diverse augmentations: spatial transformation, deformation, and additive noises. We also use adversarial training together with meta-learning.

Learning data augmentation is to train the target learner better, i.e., learning to learn better. Validation data have been used in meta-learning literatures for few-shot learning \cite{ravi2016optimization,ren2018meta,lorraine2018stochastic}, where very limited training data are available. Here we follow the MAML \cite{finn2017model} algorithm to set a meta-objective for the augmentation network. That is, augmentations conducted on a training mini-batch is evaluated on another validation one.

% The use of hard samples for data augmentation using adversarial technique has shown to improve generalization and robustness of supervised models. Virtual Adversarial Training (VAT) \cite{miyato2018virtual} searches for a minimal adversarial perturbation in the image space that changes the model’s current class label prediction. Subsequently, the model is trained with these adversarial augmentations to predict the original class. This has been shown to be effective in a semi-supervised setting and also beneficial in improving model robustness in a purely supervised setting. In a more recent work \cite{gontijo2020affinity}, the authors empirically investigate the two competing properties of the augmentations, namely, how close the data augmentation is to the true data distribution and how diverse are the augmented samples. In this work, we propose architectures and optimization schemes aimed at improving augmentation diversity with less reliance on domain knowledge while improving generalization.

%The theoretical underpinnings of data augmentation schemes are also being studied. In \cite{dao2019kernel}, the authors analyze the data augmentation under the lens of Markov process. 
\section{The Online Data Augmentation Formulation}

In this section, we introduce our online data augmentation paradigm: updating target model $\theta$ and augmentation model $\phi$ alternately. In this way, data augmentation and target model are learned jointly. Benefiting from the joint learning, the augmentation model $\phi$ can adapt to the target model $\theta$ in training. 

% It can jointly optimize data augmentation and model training. The key idea is to integrate adversarial training and meta-learning in learning data augmentation.

\begin{figure}[!t]
\setlength{\abovecaptionskip}{-0.mm}
\setlength{\belowcaptionskip}{-3.mm}
\centering
\includegraphics[width=1\linewidth]{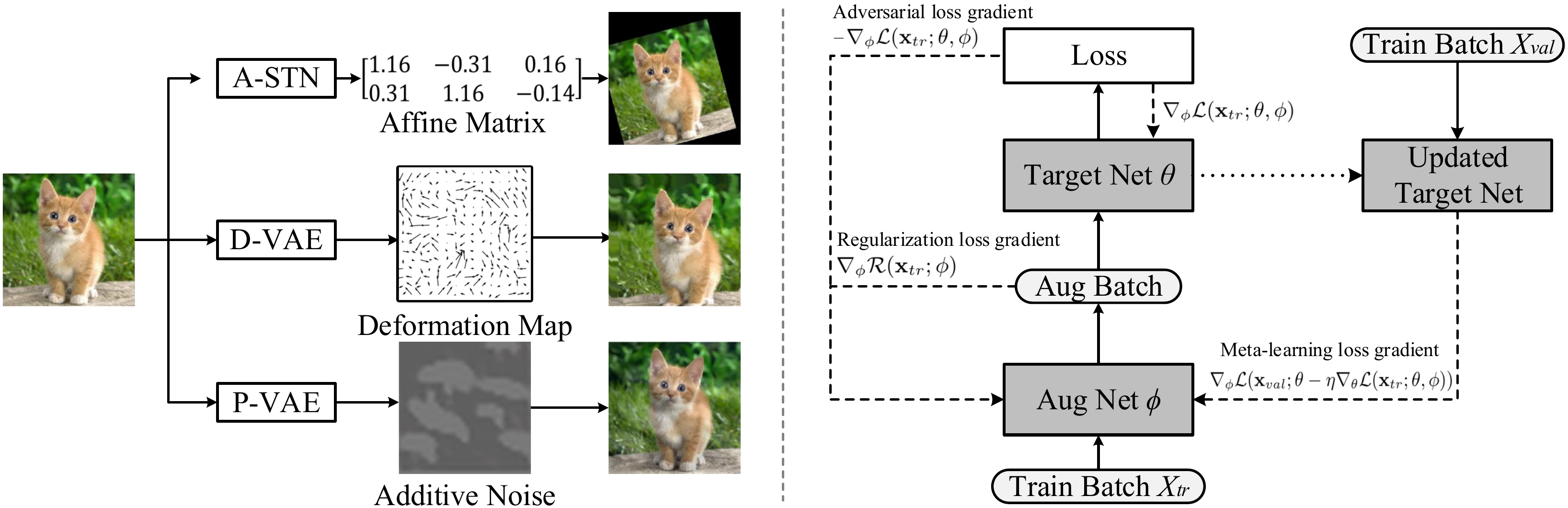}
\caption{Augmentation illustrations ({\bf Left}) and OnlineAugment scheme ({\bf Right}). We propose three models to conduct spatial transformation, deformation, and noise perturbation. OnlineAugment can jointly optimize each plug-in augmentation network with the target network. Updating the augmentation network incorporates adversarial training, meta-learning, and some novel regularizations.}
\label{fig:framework}
\end{figure}

% To this end, we draw on inspirations from adversarial training, reducing overfitting to training data, and meta-learning, explicitly modeling generalization to validation data. 

% where $\theta$ and $\phi$ are the corresponding model parameters
%  and validation set $\mathcal{D}_{val}$

For simplicity, let $\mathbf{x}$ be the annotated data, including both input and target. Note that $\mathbf{x}$ can come from any supervised task such as image classification or semantic segmentation. Let $\theta$ and $\phi$ denote the target and augmentation models. During training, the target model $\theta$ learns from the augmented data $\phi(\mathbf{x})$ instead of the original $x$. Note that $\phi$ will also transform the ground truth annotation if necessary. For example, in semantic segmentation, $\phi$ applies the same spatial transformations to both an image and its segmentation mask. Without loss of generality, we assume $\theta$ and $\phi$ are parameterized by deep neural networks, which are mostly optimized by SGD and its variants. Given a training mini-batch $\mathbf{x}_{tr}$ sampled from training set $\mathcal{D}_{tr}$, $\theta$ is updated by stochastic gradient:
\begin{equation}\label{eq:theta}
    \nabla_{\theta}
\mathcal{L}(\mathbf{x}_{tr};\theta, \phi),
\end{equation}
% Given training set $\mathcal{D}_{tr}$, $\theta$ learns from $\mathcal{D}_{tr}$ by minimizing the loss function $\mathcal{L}$ as follows:
% \begin{equation}\label{eq:theta}
%     \arg\min_{\theta}\underset{\mathbf{x}\sim \mathcal{D}_{tr}}{\mathbb{E}} \;
% [\mathcal{L}(\mathbf{x};\theta, \phi)],
% \end{equation}
where the choice of $\mathcal{L}$ depends on the task. In the case of object classification, $\mathcal{L}$ is a cross entropy function.
% minimizing the loss $\mathcal{L()}$.

% Let $(\mathbf{x}, \mathbf{y})$ be a pair of input and target. Note that $\mathbf{y}$ can be any kind of annotations such as class label or segmentation mask. Let $f_\theta$ and $f_\phi$ denote the target and augmentation models, where $\theta$ and $\phi$ are the corresponding model parameters. In standard training, given training data $\mathcal{D}_{tr}$ and validation data $\mathcal{D}_{val}$, $f_\theta$ learns from $\mathcal{D}_{tr}$ by minimizing the loss $\mathcal{L}$. In our setting, $f_\theta$ learns from the augmented data $f_\phi(\mathbf{x})$ instead of the original $x$. 

% \begin{equation}\label{eq:theta}
%     \arg\min_{\theta}\underset{\mathbf{(x,y)}\sim \mathcal{D}_{tr}}{\mathbb{E}} \;
% [\mathcal{L}(f_\theta(f_\phi(\mathbf{x})), \mathbf{y})].
% \end{equation}

% $L_{(x,y)\sim D_{tr}}(f_\theta(x), y)$. During the training, we select the best $\theta^*$ according to the validation performance on $D_{val}$.

The goal of data augmentation is to improve the generalization of the target model. To this end, we draw on inspirations from adversarial training and meta-learning. Adversarial training aims to increase the training loss of the target model by generating hard augmentations. It can effectively address the overfitting issue of the target model. Meta-learning, on the other hand, can measure the impact of augmented data on the performance of validation data. If a validation set $\mathcal{D}_{val}$ is possible, we can sample from it a validation mini-batch  $\mathbf{x}_{val}$. Otherwise, we can simulate the meta-tasks by sampling two separate mini-batches from train set $\mathcal{D}_{tr}$ as $\mathbf{x}_{tr}$ and $\mathbf{x}_{val}$. Mathematically, the stochastic gradient of augmentation model $\phi$ is computed as:
\begin{equation}\label{eq:phi}
\nabla_{\phi}\mathcal{L}(\mathbf{x}_{val};\theta-\eta\nabla_{\theta}\mathcal{L}(\mathbf{x}_{tr};\theta,\phi))+\lambda\nabla_{\phi}\mathcal{R}(\mathbf{x}_{tr};\phi)-\beta\nabla_{\phi}\mathcal{L}(\mathbf{x}_{tr};\theta,\phi),
\end{equation}
% Instead of using additional validation data, we sample independent mini-batches $\mathbf{x}_1$ and $\mathbf{x}_2$ from training data to form the meta-tasks. Mathematically, the objective of learning the augmentation model $\phi$ is:
% \begin{equation}\label{eq:phi}
% \arg\min_{\phi} \; \underset{\mathbf{x}_2\sim \mathcal{D}_{tr}}{\mathbb{E}} \;
% [\mathcal{L}(\mathbf{x}_2;\theta(\phi))]+\underset{\mathbf{x}_1\sim \mathcal{D}_{tr}}{\mathbb{E}} \;
% [\lambda\mathcal{R}(\mathbf{x}_1;\phi)-\beta\mathcal{L}(\mathbf{x}_1;\theta,\phi)],
% \end{equation}
% \begin{equation}\label{eq:phi}
% \arg\min_{\phi} \; \underset{\mathbf{(x,y)}\sim \mathcal{D}_{val}}{\mathbb{E}} \;
% [\mathcal{L}(f_\theta(\mathbf{x}), \mathbf{y})]-\underset{\mathbf{(x,y)}\sim \mathcal{D}_{tr}}{\mathbb{E}} \;
% [\mathcal{L}(f_\theta(f_\phi(\mathbf{x})), \mathbf{y})].
% \end{equation}
% \begin{equation}\label{eq:phi}
% \arg\min_{\phi} \; \left(\underset{\mathbf{x}\sim \mathcal{D}_{tr}}{\mathbb{E}} \;
% [\lambda\mathcal{R}(\mathbf{x};\phi)-\beta\mathcal{L}(\mathbf{x};\theta(\phi))] + \arg\min_{\theta} \; \underset{\mathbf{x}\sim \mathcal{D}_{tr}}{\mathbb{E}} \;
% [\mathcal{L}(\mathbf{x};\theta(\phi))]\right),
% \end{equation}
% That is, $\theta$ updated by augmented data $\phi(\mathbf{x}_1)$ is further evaluated on $\mathbf{x}_2$ to update $\phi$.
where $\mathcal{L}(\mathbf{x}_{val};\theta-\eta\nabla_{\theta}\mathcal{L}(\mathbf{x}_{tr};\theta,\phi))$, $\mathcal{R}(\mathbf{x}_{tr};\phi)$, and $-\mathcal{L}(\mathbf{x}_{tr};\theta,\phi)$ are the generalization, regularization, and adversarial losses. $\lambda$ and $\beta$ are the balancing weights. $\theta-\eta\nabla_{\theta}\mathcal{L}(\mathbf{x}_{tr};\theta,\phi)$ represents the the updated target network by augmented data $\phi(\mathbf{x}_{tr})$. For simplicity, here we use a vanilla gradient descent with learning rate $\eta$. Other more complex optimizers are also applicable. For efficient training, we use the second-order approximation \cite{liu2018darts} to compute the meta-gradient.

% $\theta(\phi)$ means the updated target network by augmented data $\phi(\mathbf{x}_1)$. We evaluate $\theta(\phi)$ on clean data $\mathbf{x}_2$ to propagate gradients to augmentation model $\phi$. We use the second-order approximation \cite{} to compute the meta-graidents. 

$\mathcal{R}(\mathbf{x};\phi)$ measures the distance between the original and augmented data. Adding this regularization term is to constrain the augmented data within reasonable distributions. Otherwise, adversarial training may cause meaningless augmentations that hurt training. Theoretically, the generalization term can also help regularize the augmentations implicitly. In practice, we find that the explicit regularization term is critical for practical adversarial training. Besides, adversarial training is performed by minimizing the negative training loss  $-\mathcal{L}(\mathbf{x}_{tr};\theta,\phi)$. In this way, the augmentation model $\phi$ learn to generate hard augmentations. The training scheme is presented in Algorithm \ref{alg:OnlineAugment} and the right figure in Fig. \ref{fig:framework}.

\begin{algorithm}[t!]
\SetAlgoLined
\KwIn{Initial target model $\theta$, initial augmentation model $\phi$, training set $\mathcal{D}_{tr}$,  and validation set $\mathcal{D}_{val}$}
\While{not converged}{

% Sample mini-batches of training and validation data, augment the training mini-batch by $f_\phi$, and compute the loss $\mathcal{L}_{train}$

Sample mini-batches $\mathbf{x}_{tr}$ and $\mathbf{x}_{val}$ from $\mathcal{D}_{tr}$ and $\mathcal{D}_{val}$ respectively 

Update augmentation model $\phi$ by stochastic gradient: $\nabla_{\phi}\mathcal{L}(\mathbf{x}_{val};\theta-\eta\nabla_{\theta}\mathcal{L}(\mathbf{x}_{tr};\theta,\phi))+\lambda\nabla_{\phi}\mathcal{R}(\mathbf{x}_{tr};\phi)-\beta\nabla_{\phi}\mathcal{L}(\mathbf{x}_{tr};\theta,\phi)$

% Generate new augmentation mini-batches using the updated augmentation model $f_\phi$ and compute the new loss $\mathcal{L}_{train}$

Update target model $\theta$ by stochastic gradient: $\nabla_{\theta}\mathcal{L}(\mathbf{x}_{tr};\theta,\phi)$
}
\KwOut{Optimized target model $\theta^*$}
\caption{OnlineAugment: Online Data Augmentation} \label{alg:OnlineAugment}
\end{algorithm}

% \begin{equation}
%     \nabla_{\theta_T}T(\cdot)=\sum_{t'=1}^{K}\nabla_{\theta_T}T(\hat{x}_{t'}).
% \end{equation}
{\bf Relation to offline augmentation methods.} 
The formulation of our online augmentation differs from the previous offline ones \cite{cubuk2019autoaugment,cubuk2019randaugment,ho2019population} mainly in three aspects. First, OnlineAugment alternates updating the target model $\theta$ and augmentation model $\phi$. The offline methods usually perform a full optimization for $\theta$ in each step of updating $\phi$. Second, to get the optimized target model, the offline methods usually require a two-stage training: learning policies and applying them. However, OnlineAugment can optimize the target model in one training process. Third, we use adversarial training in learning the augmentation model. The offline methods only have one generalization objective, maximizing performance on validation data.

{\bf Relation to adversarial training methods.} 
Adversarial training \cite{goodfellow2014explaining,kurakin2016adversarial,madry2017towards} is mainly used to improve the robustness of the target model. Some works \cite{madry2017towards,xie2019feature} have shown that robust models usually come at the cost of degraded generalization to clean data. The goal of OnlineAugment is generalization rather than robustness. To pilot adversarial training for generalization, we design new regularization terms and add meta-learning in OnlineAugment. 
% The formulation of our online augmentation differs from the previous offline ones \cite{cubuk2019autoaugment,cubuk2019randaugment,ho2019population} mainly in three aspects. First, we add the adversarial training objective in optimizing the data augmentation model. In contrast, the offline methods only have the generalization objective. Second, we use efficient online training that alternates updating the target model and data augmentation. The offline ones treats optimizing the target model as an inner optimization in updating the data augmentation model. That is, every step of updating $\phi$, they perform a full optimization for $\theta$. It is also the main reason causing low efficiency for offline methods. Finally, we can get the final target model in one training round using Algorithm \ref{alg:OnlineAugment}. However, the offline methods require two separated training stages for the data augmentation and the final target models.

\begin{figure}[!t]
\centering
\includegraphics[width=1\linewidth]{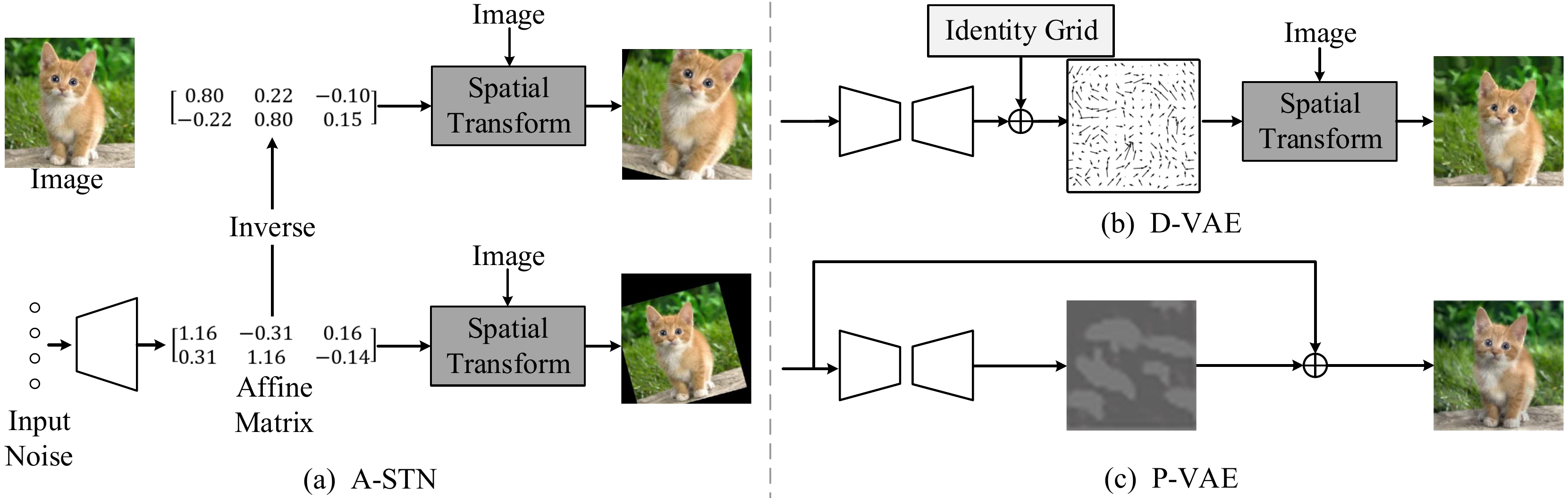}
\caption{Augmentation models: A-STN {\bf(a)}, D-VAE {\bf(b)}, and P-VAE {\bf(c)}. A-STN, conditioned on Gaussian noise, generate a transformation matrix. Both the matrix and its inverse are applied to an image for diversity. D-VAE or P-VAE takes an image as input,  generating deformation grid maps or additive noise maps. The three models are trainable if plugged in the training scheme in Figure \ref{fig:framework}.}
\label{fig:augnet}
\end{figure}

\section{Data Augmentation Models}
After introducing the OnlineAugment scheme, we present three different data augmentation models in this section. The three models are motivated by our analysis of possible data transformations.  Specifically, we summarize them into three types: global spatial transformation, local deformation, and intensity perturbation. Each transformation corresponds to a model below. They either change pixel locations or values. Note that the pixel in this work refers to an element in a generic feature map, not necessarily an image. Technically, we design the augmentation models based on the spatial transformer network (STN) \cite{jaderberg2015spatial} and variational auto-encoder (VAE) \cite{kingma2013auto}.

\subsection{Global Spatial Transformation Model}
There are several commonly used spatial transformation functions such as rotation, scale, and translation, which can be unified into more general ones, such as affine transformation. It is well-known that STN \cite{jaderberg2015spatial} can perform general spatial transformations. Briefly, STN contains three parts: a localization network, a grid generator, and a sampler. The last two modules are deterministic and differentiable functions, denoted as $\tau$. Therefore, our focus is to design a new localization network $\phi$ that outputs the transformation matrix. 

% {\bf Augmentation STN.} Suppose we use affine transformation, the output should be a 6-dimension vector, further re-shaped into a 2$\times$3 transformation matrix. Traditionally, the STN localization network uses images or feature maps as inputs. However, in our setting, we find they make training the localization network unstable. Because the STN plays the opposite role here. The original STN is to reduce the input variance for the target network. On the contrary, the goal of data augmentation is to increase the input variance, which makes training the localization network more difficult. Directly mapping the complex distributions of images or feature maps into some proper 6-dimension subspace is nontrivial. Small perturbations in the transformation space may lead to large differences between transformed data. Besides, our observations also indicate that, for the purpose of data augmentation, the transformations are transferable between input samples. That is, the same transformation is helpful if applied to different inputs.

{\bf Augmentation STN (A-STN).} Suppose we use affine transformation, the output should be a 6-dimension vector, further re-shaped into a 2$\times$3 transformation matrix. Traditionally, the STN localization network uses images or feature maps as its input. Here we also provide an alternative input of Gaussian noises. Our design is motivated by the observation that the global spatial transformations are transferable in data augmentation. That is, the same spatial transformation is applicable to different images or feature maps in augmenting training data. Therefore, conditioning on the images or feature maps may not be necessary for generating the augmentation transformation. 

The architecture of the localization network $\phi$ depends on the choices of its input. It can be a convolutional neural network (CNN) or a multi-layer perceptron (MLP) if conditioned on the generic feature map or 1-D Gaussian noise. We will give its detailed architectures in the experiment. Moreover, the MLP localization network itself is also transferable as it is unrelated to the target task. However, we may need to craft new CNN localization networks for different tasks. Therefore, it is preferable to implement the localization network $\phi$ as an MLP in practice.

% Based on above analysis, we propose to use Gaussian noises as the input of the localization network. Gaussian distribution is simple, resulting in smooth training for the localization network. Gaussian noises are also independent of the input images or features. In principle, the localization network learns to map Gaussian distribution to some transformation subspace, which is shared by all the training data. Since both the input and output can be 1-dimensional, we design the localization network as a simple multi-layer perceptron (MLP). We will give its detailed architecture in the experimental section.

% Suppose the localization network produces the spatial transformation $\tau_{\phi}$

{\bf Double Cycle-consistency Regularization.} To apply the spatial transformation model to Algorithm \ref{alg:OnlineAugment}, we need to design a proper regularization term $\mathcal{R}$. Empirically, increasing the spatial variance of training data can enhance the generalization power of the model. However, excessive spatial transformations probably bring negative impacts. To constrain the spatial transformations within reasonable scopes, we propose a novel double cycle-consistency regularization, see in Fig. \ref{fig:augnet} (a). The key idea is to measure the lost information during the spatial transformation process. Mathematically, we compute the double cycle-consistency loss:
\begin{equation}
    \mathcal{R}_c^s(\mathbf{x};\phi)=\|\tau_{\phi}^{-1}(\tau_{\phi}(\mathbf{x}))-\mathbf{x}\|_2^2+\|\tau_{\phi}(\tau_{\phi}^{-1}(\mathbf{x}))-\mathbf{x}\|_2^2,
\end{equation}
where $\tau_{\phi}(\mathbf{x})=\tau(\mathbf{x},\phi(\mathbf{z}))$. The deterministic function $\tau$ transforms the image or feature map $\mathbf{x}$ using the generated affine matrix $\phi(\mathbf{z})$, where $\mathbf{z}$ is the Gaussian noise. $\tau_{\phi}^{-1}$ denotes the inverse transformation of $\tau_{\phi}$. Ideally, applying a transformation followed by its inverse will recover the original input, and vice versa. In reality, whichever applied first may cause some irreversible information loss. For example, the zoom-in transformation discards the region falling out of scope. Applying the zoom-out transformation afterwards will produce zero or boundary padding, which is different from the original image region. We find a single cycle-consistency loss will lead to biased transformations. The localization network $\phi$ tends to output zoom-out transformations whose inverse can easily recover the original input. Fortunately, imposing the double cycle-consistency constraint can avoid the biases effectively, thereby producing more diverse transformations.

\subsection{Local Deformation Model}
Apart from the global spatial transformation model, we propose another complementary deformation model. The global transformation applies the same transformation matrix to all the pixels of an image or feature map. In the local deformation, each pixel, however, has an independent transformation. 

% Note that the pixel refers to an element in a generic feature map, not necessarily an image. 

{\bf Input and Output.} It is cumbersome and also unnecessary to produce all the transformation matrices. Recall that STN performs transformations by the grid sampling. A better choice is to predict a grid map directly. For 2D transformations, the grid map has the shape $h\times w\times 2$, where $h$ and $w$ are the height and width of the input feature map. Each location in the grid map indicates the 2D coordinates to sample a pixel. A grid map is personalized to an image or feature map as each pixel has its own transformation. Different from a low-dimension affine matrix, a deformation grid map may be unlikely to transfer. Therefore, our deformation model is conditioned on the image or feature map, generating the grid map.

% {\bf Input.} As each pixel has its own transformation, a grid map is personalized to an input image or feature map. Different from a low-dimension affine matrix, a deformation grid map is unlikely to transfer from one input to another. Therefore, our deformation model is conditioned on the image or feature map, generating the grid map. Compared with the affine transformation, the local deformations can bring more consistent transformations. More specifically, small changes in the grid maps result in similar transformed inputs. Benefiting from this property, using the image or feature map as input raise no instability in training the deformation model.

{\bf Deformation VAE (D-VAE).} The deformation model $\phi$, see in Fig. \ref{fig:augnet} (b),  builds on the Variational Autoencoders (VAE) \cite{kingma2013auto}, a popular generative model. A VAE model consists of an encoder and a decoder. Similar to the original VAE, our deformation VAE also uses images or feature maps as the encoder input. However, in our setting, the decoder outputs the deformation grid maps instead of the reconstructed input. We refer to the deformation grid maps as deformation deltas $\Delta_\phi^d$. They are added on the grid maps of identity mapping $id$ to perform grid sampling $\tau$. The transformed input $\tau(\mathbf{x},\Delta_\phi^d+id)$ serves as the reconstructed input. Following the original VAE, our deformation VAE is also trained to minimize both the reconstruction loss and the KL-divergence between the encoded distribution and the standard normal distribution:
% \begin{equation}
%     \mathcal{R}_v^d(\mathbf{x};\phi)=\|\mathbf{x}-\phi(\mathbf{x})\|_2^2+\mathcal{KL}(\mathcal{N}(\mu_\mathbf{x}, \Sigma_\mathbf{x}), \mathcal{N}(0, I)),
% \end{equation}

%\begin{equation}\label{eq:deform-vae}
%    \mathcal{R}_v^d(\mathbf{x};\phi)=\|\mathbf{x}-%\tau(\mathbf{x},\Delta_\phi^d+id)\|_2^2+\mathcal{K%L}(\mathcal{N}(\mu_\mathbf{x}, \Sigma_\mathbf{x}), %\mathcal{N}(0, I)),
%\end{equation}

\begin{equation}\label{eq:deform-vae}
    \mathcal{R}_v^d(\mathbf{x};\phi)=\|\mathbf{x}-\tau(\mathbf{x},\Delta_\phi^d+id)\|_2^2+\mathcal{KL}(\mathcal{N}(\mu_{\phi}(\mathbf{x}), \Sigma_{\phi}(\mathbf{x})), \mathcal{N}(0, I)),
\end{equation}

where $\mu_{\phi}(\mathbf{x})$ and $\Sigma_{\phi}(\mathbf{x})$ are the encoded mean and variance, parameterizing the Gaussian distribution.

% (\prasanna{Changed $\mu_\mathbf{x}$ to $\mu(\mathbf{x})$})

% In our setting, the VAE, conditioned on the image or feature map, outputs the deformation grid maps, referred to as deformation deltas. The final grid maps adds up the deformation deltas and the identity mapping grid maps. Following 

% We propose a residual VAE as our deformation model. The residual VAE, conditioned on the image or feature map, outputs the deformation grid map. 

{\bf Smoothness Regularization.} Smooth deformations are essential to preserving the quality of deformed data. Otherwise, the deformed data may become noisy as each pixel is sampled from an independent location in the original image or feature map. This is especially important for location-sensitive tasks such as semantic segmentation. Given an arbitrary pixel $i$ and its neighbours $j\in \mathcal{N}(i)$, we enforce the local smoothness constraint on the deformation deltas:
\begin{equation}
    \mathcal{R}_s^d(\mathbf{x};\phi)=\sum_i\sum_{j\in \mathcal{N}(i)}\|\Delta_\phi^d(i)-\Delta_\phi^d(j)\|_2^2.
\end{equation}
The smoothness regularization can make the deformations consistent for nearby pixels. In the experiment, we use the combination $\lambda_v^d\mathcal{R}_v^d(\mathbf{x};\phi)+\lambda_s^d\mathcal{R}_s^d(\mathbf{x};\phi)$ to regularize our deformation augmentation model.

\subsection{Intensity Perturbation Model}
The above two models perform data augmentation by changing the pixel locations. Here we propose another model to manipulate the pixel values instead. As an analogy, the offline data augmentation methods \cite{cubuk2019autoaugment,cubuk2019randaugment,ho2019population} use some built-in image texture processing functions such as colour and brightness. These functions are designed based on the domain knowledge for natural images. In contrast, our intensity perturbation is more general without domain knowledge.

{\bf Perturbation VAE (P-VAE).} Specifically, the intensity perturbation model $\phi$, see in Fig. \ref{fig:augnet} (c), conditioned on the image or feature map, generates additive noises $\Delta_\phi^p$. As the deformation, we use the VAE model to learn the intensity perturbation. The reconstructed input is the sum of the input and generated noises $\mathbf{x}+\Delta_\phi^p$. Therefore, we can compute the VAE loss as:
\begin{equation}\label{eq:perturb-vae}
    \mathcal{R}_v^p(\mathbf{x};\phi)=\|\Delta_\phi^p\|_2^2+\mathcal{KL}(\mathcal{N}(\mu_{\phi}(\mathbf{x}), \Sigma_{\phi}(\mathbf{x})), \mathcal{N}(0, I)),
\end{equation}

where $\mu_{\phi}(\mathbf{x})$ and $\Sigma_{\phi}(\mathbf{x})$ are the mean and variance of the encoded Gaussian distribution. Note that P-VAE produces deltas $\Delta_\phi^p$ in the image or feature map domain while the deltas $\Delta_\phi^d$, predicted by D-VAE, lie in the grid map domain. It results in the different reconstruction losses in Equations \ref{eq:perturb-vae} and \ref{eq:deform-vae}.

{\bf Relation to Adversarial Attacks.} Additive noise is a common tool in generating adversarial examples \cite{xie2019adversarial,madry2017towards}. Here we explore its potential in data augmentation. Although the adversarial examples serve as augmented data in adversarial training, they are mainly to improve the model's robustness. Some evidence \cite{raghunathan2019adversarial} have shown that adversarial training usually sacrifices the model generalization to clean data. However, our intensity perturbation model can improve the generalization through the OnlineAugment. Recently, concurrent work AdvProp \cite{xie2019adversarial} successfully adapts the PGD attack \cite{madry2017towards} to data augmentation. In contrast, we design the perturbation VAE model to generate additive noises.

% We name it noisy perturbation as an analogy to the texture functions in image processing such as color and brightness. However, our texture perturbation is more general. The image processing functions make structural changes in image texture. Each function is designed to change one image property. For instance, if we enhance 

% \input{implementation}
\section{Experiments}
In this section, we empirically evaluate OnlineAugment with the three models.
% In this section, we empirically evaluate the OnlineAugment scheme. More specifically, we plug in the three augmentation models: Augmentation STN (A-STN), Deformation VAE (D-VAE), and Perturbation VAE (P-VAE). The evaluation is conducted on two tasks: image classification and medical image segmentation.

\subsection{Experimental Settings}
Applying OnlineAugment is simple in practice. The augmentation models A-STN, D-VAE, and P-VAE requires neither pre-training \cite{lee2019learning} nor early stopping \cite{ratner2017learning}. Because they can adapt to the target network during online training. Inspired by AdvProp \cite{xie2019adversarial}, we use multiple batch normalization layers to merge different types of augmentations. A-STN, D-VAE, and P-VAE are trained by Adam optimizer with the same learning rate of $1e-3$, weight decay $1e-4$, and $\beta_1=0.5$. Other Adam hyper-parameters are set by default in Pytorch.

{\bf A-STN.} We design the noise conditioned A-STN as a 6-layer MLP. Specifically, it takes only 1-dimensional Gaussian noises as input and outputs 6-dimensional affine parameters. Each hidden layer generates 8-dimension features. Batch normalization, ReLU, and dropout (0.5) are applied after each linear layer. The loss weights are set as $\lambda^s_c=0.1$ and $\beta^s=0.1$. 

{\bf D-VAE.} D-VAE consists of an encoder and a decoder. The encoder maps an image to the 32-dimensional latent space. It includes 5 $3\times 3$ convolutional layers and three linear layers. The first convolutional layer increases the channel number from 3 to 32. After that, the feature channels double if the convolution stride is 2. There are two convolutional layers on each resolution. The first linear layer takes the reshaped convolutional features and outputs 512-dimensional latent features. Another two linear layers generate the mean and variance vectors of encoded Gaussian distributions. The decoder is simply the reverse of the encoder with transposed convolutions. The last layer is a $1\times 1$ convolution producing 2-channel grid maps. We use the weights $\lambda^d_v=1$, $\lambda^d_s=10$, and $\beta^d=1e-2$.

{\bf P-VAE.} It shares almost the same architecture as D-VAE. The only difference is the last layer in the decoder, because P-VAE needs to generate additive noises on the images. The latent space dimension is set to 8 for P-VAE. We also set the hyper-parameters $\lambda^p_v=1e-3$ and $\beta^p=10$.

{\bf Image Classification.} We use datasets CIFAR-10, CIFAR-100, SVHN, and ImageNet with their standard training/test splits. To tune hyper-parameters efficiently, we also sample reduced datasets: R-CIFAR-10, R-CIFAR-100, and R-CIFAR-SVHN. Each of them consists of 4000 examples. The sampling is reproducible using the public sklearn StratifiedShuffleSplit function with random state 0. We report top-1 classification accuracy in most tables except for Table \ref{tb:sota} with top-1 errors. The target networks are Wide-ResNet-28-10 \cite{zagoruyko2016wide} (W-ResNet), Shake-Shake network \cite{gastaldi2017shake}, and ResNet-50 \cite{he2016deep}. We use Cutout \cite{devries2017improved} as the baselines in Tables \ref{tb:ablation-a-stn}, \ref{tb:ablation-d-vae}, \ref{tb:ablation-p-vae}, and \ref{tb:ours-vs-aa}.

{\bf Medical Image Segmentation.}
To test the generalization ability of our approach, we further conduct experiments on the medical image segmentation dataset LiTS \cite{bilic2019liver}. LiTS published 131 liver CT images as a training set with liver and tumor annotations. Since these experiments are to prove the effectiveness of our augmentation algorithm, pursuing the highest performance is not our goal, we use the 2D UNet as the segmentation network to segment CT images on the axial plane slice by slice. We randomly split the 131 CT images into 81 training cases, 25 validation cases, and 25 test cases. We first resample all images to $2\times2\times2$ mm, then center crop each slice to $256\time256$ in size, and finally normalize the image window [-200, 250] to [0, 1] as the input of the network. Only A-STN and D-VAE are presented in this task, since P-VAE has no obvious performance improvement. Compared with classification tasks, segmentation tasks are sensitive to location. Therefore, for A-STN and D-VAE, we not only perform a bilinear grid sample on the images, but also perform a nearest neighbor grid sample for the label masks using the same grid.

We also compared our proposed OnlineAugment with RandAugment \cite{cubuk2019randaugment}. We slightly modify RandAugment to fit our setting. As the number of transformations involved is relatively small, all transformations are used during training. Therefore, for global spatial transformation, the search space is the magnitude of 4 transforms, including rotate, translate-x, translate-y, and scale. For the deformation model, the search space is the magnitude of local deformations. At each iteration, the magnitude of each transformation is uniformly and randomly sampled between 0 and the upper bound for both global spatial transformation and local deformation. 

\begin{table}[t]
\begin{center}
\caption{Evaluation of the Gaussian noise input and double cycle-consistency regularization in A-STN. We compare them to the image condition and single cycle-consistency. Double cycle-consistency outperforms the single one. With the double one, the noise and image inputs get comparable accuracy.}\label{tb:ablation-a-stn}
% \small
% \setlength\tabcolsep{1.5pt}
% {@{}lcccccccc@{}}
\begin{tabular}{llccccc}
\toprule
Dataset & Model & Cutout & Image Input & Image  Input & Noise Input & Noise Input \\
& & & +1 cycle & +2 cycles & +1 cycle & +2 cycles\\
\hline
R-CIFAR-10 & W-ResNet & 80.95 & 83.24 & 84.76 & 82.62 & {\bf 84.94}\\
% 84.24
% \multicolumn{7}{c}{{\bf Reduced CIFAR-10}}\\

\bottomrule
\end{tabular}
\end{center}
% \vspace{-2em}
% \vspace{-2mm}
\end{table}

\begin{table}[t]
\setlength{\abovecaptionskip}{-0.mm}
\setlength{\belowcaptionskip}{-4.mm}
\begin{center}
\caption{Evaluation of the smoothness regularization (SR) in D-VAE. We report the results on both image classification and segmentation. The smoothness regularization is more useful for the location-sensitive image segmentation task. }\label{tb:ablation-d-vae}
% \small
\setlength\tabcolsep{3pt}
% {@{}lcccccccc@{}}
\begin{tabular}{llccc}
\toprule
Dataset & Model & Baseline & D-VAE Only & D-VAE + SR \\
\hline
R-CIFAR-10 & W-ResNet & 80.95 (Cutout) & 82.72 & {\bf 82.86}\\
Liver Segmentation & U-Net & 66.0 (No Aug.) & 68.51 & {\bf 70.49}\\
\bottomrule
\end{tabular}
\end{center}
% \vspace{-1em}
% \vspace{-4mm}
\end{table}

\subsection{Experimental Results}
%  We compare them to the image condition and single cycle-consistency.

% The experiments consist of five parts. First, we conduct ablation studies for ASTN, D-VAE, and P-VAE. Then we not only compare them with AutoAugment \cite{cubuk2019autoaugment} but also study their orthogonality. The comparisons with other state-of-theart data augmentation methods are also reported. Last but not least, we also
% evaluate OnlineAugment in a medical segmentation task.

The experiments consist of ablation studies for the three models, comparisons with AutoAugment \cite{cubuk2019autoaugment}, and their orthogonality. The comparisons with state-of-the-art methods are reported on image classification and medical segmentation.

% \vspace{-0.2cm}
\subsubsection{A-STN.} The A-STN may be conditioned on image or Gaussian noise. We compare these two choices in this ablation study. Besides, we also compare its regularization with single or double cycle-consistency losses. Table \ref{tb:ablation-a-stn} gives the comparisons. The double cycle-consistency obtains higher accuracy than the single cycle one. Because it can make A-STN produce more diverse transformations. With the double-cycle consistency regularization, the noise and image conditions achieve comparable accuracy. We use the noise condition A-STN in other experiments since its architecture is transferable between tasks.
% \vspace{-0.4cm}
\subsubsection{D-VAE.} The smoothness regularization comes as an additional component in the D-VAE. We evaluate its effectiveness in both CIFAR-10  classification and liver segmentation tasks. Table \ref{tb:ablation-d-vae} presents the results. Interestingly, the smoothness regularization has little effect on classification accuracy. However, it makes a difference (\%2) for the liver segmentation. Because the liver segmentation is a location-sensitive task. The smoothness regularization can remove the noises along the boundaries of segmentation masks. 
% \vspace{-0.4cm}
\subsubsection{P-VAE.} The P-VAE generates additive noises to perform data augmentation. AdvProp \cite{raghunathan2019adversarial} has a similar goal, but utilizes the iterative gradient methods for the generation. To make adversarial training helpful for the generalization to clean data, AdvProp requires to set proper step length and number in iterative gradient methods. For a fair comparison, we use only adversarial training and the noise regularization in training P-VAE. Table \ref{tb:ablation-p-vae} shows the comparisons. P-VAE compares favorably to AdvProp with GD and I-FGSM. On the one hand, P-VAE learns some noise distributions while the iterative methods rely on the deterministic gradient ascent rules. On the other hand, P-VAE generates structured noises, while the iterative approaches produce more complex ones.

% The P-VAE outperforms AdvProp with three different iterative gradient methods: PGD, GD, and I-FGSM. This can be attributed to our OnlineAugment formulation and the VAE based generation. First, OnlineAugment integrates both adversarial training and also meta-learning aimed at better generalization, while AdvProp uses adversarial training only. Second, the P-VAE models the distribution of noises. The gradient-based generation is deterministic given the target network state.

% PGD \cite{madry2017towards}, GD, and I-FGSM \cite{chang2018efficient}. We plug them in state-of-the-art adversarial augmentation
\begin{table}[t]
\setlength{\abovecaptionskip}{-0.mm}
\setlength{\belowcaptionskip}{-0.mm}
\begin{center}
\caption{Comparisons of P-VAE to AdvProp \cite{xie2019adversarial} with iterative gradient methods PGD \cite{madry2017towards}, GD, and I-FGSM \cite{chang2018efficient}. Adversarial training plus only the noise regularization can make P-VAE comparable to AdvProp with GD or I-FGSM.}\label{tb:ablation-p-vae}
% \small
\setlength\tabcolsep{5pt}
% {@{}lcccccccc@{}}
\begin{tabular}{llccccc}
\toprule
Dataset & Model & Cutout & AP+PGD & AP+GD & AP+I-FGSM & P-VAE\\
\hline
R-CIFAR-10 & W-ResNet & 80.95 & 83.00 & 83.90 & 83.92 & {\bf 84.08} \\
% \multicolumn{7}{c}{{\bf Reduced CIFAR-10}}\\
\bottomrule
\end{tabular}
\end{center}
% \vspace{-2em}
% \vspace{-2mm}
\end{table}

\begin{table}[t]
\setlength{\abovecaptionskip}{-0.mm}
\setlength{\belowcaptionskip}{-4.mm}
\begin{center}
% A-STN, D-VAE, and P-VAE can separately make improvements on top of Cutout.
\caption{Ours {\it v.s.} AutoAugment (AA). The three models helps separately, and they together may perform on a par with AA. The stochastic shake-shake operations may interfere with the online learning, reducing the improvements. }\label{tb:ours-vs-aa}
% \small
% \setlength\tabcolsep{1.5pt}
% {@{}lcccccccc@{}}
\begin{tabular}{llcccccc}
\toprule
Dataset & Model & Cutout & AA & A-STN & D-VAE & P-VAE & Comb.\\
\hline
% \multicolumn{7}{c}{{\bf Reduced CIFAR-10}}\\
\multirow{2}{*}{R-CIFAR-10} & Wide-ResNet-28-10 & 80.95 & 85.43 & 84.94 & 82.72 & 84.18 & {\bf 85.65}\\
& Shake-Shake (26 2x32d) & 85.42 & {\bf 87.71} & 86.62 & 86.51 & 86.34 & 87.12\\
% & 81.56 & 87.71 & 83.13 & 83.04 & 83.45 & 83.76\\
\hline 
\multirow{2}{*}{R-CIFAR-100} & Wide-ResNet-28-10 & 41.64 & 47.87 & 46.55 & 45.42 & 47.45 & {\bf 48.31}\\
& Shake-Shake (26 2x32d) & 44.41 & {\bf 48.18} & 46.81 & 46.53 & 46.30 & 47.27\\
% \multirow{2}{*}{{\bf R. CIFAR-100}} \\
\hline 
\multirow{2}{*}{R-SVHN} & Wide-ResNet-28-10 & 90.16 & 93.27 & 92.73 & 91.32 & 91.61 & {\bf 93.29}\\
& Shake-Shake (26 2x32d) & 94.03 & {\bf 94.63} & 94.15 & 94.06 & 94.12 & 94.21 \\
\bottomrule
\end{tabular}
\end{center}
% \vspace{-2em}
% \vspace{-8mm}
\end{table}

{\bf OnlineAugment {\it v.s.} AutoAugment.} AutoAugment is a representative offline augmentation method. Here we evaluate OnlineAugment by comparing with it. Table \ref{tb:ours-vs-aa} provides the separate results of three data augmentation models, as well as their combined. Each model, independently, can boost the generalization of two target networks on three datasets. Combining them can bring further improvements, achieving comparable performance as AutoAugment. We can also observe that the improvements for the Shake-Shake network \cite{gastaldi2017shake} are lower than those of the Wide ResNet \cite{zagoruyko2016wide}. One possible explanation is that the stochastic shake-shake operations may affect the online learning of data augmentation.

% Because the Shake-Shake network itself has some augmentation operations inside.
% of three independent models and also their combination.
{\bf OnlineAugment + AutoAugment.} 
Apart from comparing OnlineAugment with AutoAugment, it is more interesting to investigate their orthogonality. Table \ref{tb:ours+aa} summarizes the results. We can find that OnlineAugment can bring consistent improvements on top of AutoAugment for different target networks and datasets. Their orthogonality comes from the differences in training schemes and augmentation models. Different from OnlineAugment, AutoAugment learns data augmentation policies in an offline manner. Moreover, the three models (A-STN, D-VAE, and P-VAE) generate different augmentations from the image processing functions in AutoAugment. Note that OnlineAugment is also orthogonal to other offline methods \cite{cubuk2019autoaugment,cubuk2019randaugment,ho2019population} since they have similar policies as AutoAugment.

{\bf Comparisons with State-of-the-art Methods.} Besides AutoAugment, we also compare OnlineAugment with other state-of-the-art methods such as Fast AutoAugment (FAA) and Population-based Augmentation (PBA). They all belong to offline data augmentation. OnlineAugment alone gets the lowest test errors on CIFAR-10 and CIFAR-100 and comparable errors on ImageNet. Further, OnlineAugment, together with AutoAugment, achieves new state-of-the-art results on all the three image classification datasets. 

% We leave the combinations of OnlineAugment with other offline methods for future investigations.

\begin{table}[t]
\begin{center}
\caption{Ours+AutoAugment (AA). We use A-STN, D-VAE, and P-VAE on top of the AutoAugment polices. Surprisingly, each model can further improve AutoAugment performance. It demonstrates that OnlineAugment is orthogonal to AutoAugment. The three models use more general augmentation operations. }\label{tb:ours+aa}
% \small
% \setlength\tabcolsep{1.5pt}
% {@{}lcccccccc@{}}
\begin{tabular}{llc|cccc}
\toprule
Dataset & Model & AA & +A-STN & +D-VAE & +P-VAE & +Comb.\\
\hline
% \multicolumn{7}{c}{{\bf Reduced CIFAR-10}}\\
\multirow{2}{*}{R-CIFAR-10} & Wide-ResNet-28-10 & 85.43 & 89.39 & 87.40 & 87.63 & {\bf 89.40}\\
& Shake-Shake (26 2x32d) & 87.71 & 89.25 & 88.43 & 88.52 & {\bf 89.50}\\
\hline 
\multirow{2}{*}{R-CIFAR-100} & Wide-ResNet-28-10 & 47.87 & 52.94 & 50.01 & 51.02 & {\bf 53.72}\\
& Shake-Shake (26 2x32d) & 48.18 & {\bf 50.58} & 50.42 & 50.87 & 50.11\\
% \multirow{2}{*}{{\bf R. CIFAR-100}} \\
\hline 
\multirow{2}{*}{R-SVHN} & Wide-ResNet-28-10 & 93.27 & 94.32 & 93.72 & 94.17 & {\bf 94.69}\\
& Shake-Shake (26 2x32d) & 94.63 & 95.21 & 94.87 & {\bf 95.28} & 95.06\\
\bottomrule
\end{tabular}
\end{center}
% \vspace{-2em}
% \vspace{-2mm}
\end{table}

\begin{table}[t]
\setlength{\abovecaptionskip}{-0.mm}
\setlength{\belowcaptionskip}{-2.mm}
\begin{center}
\caption{Comparisons with state-of-the-art methods. We compare our OnlineAugment with AutoAugment (AA), PBA \cite{ho2019population}, and Fast AutoAugment (FAA) \cite{lim2019fast} on three datasets. OnlineAugment alone obtains comparable test errors. Combining it with AutoAugment produces the lowest errors on three datasets. }\label{tb:sota}
% \small
% \setlength\tabcolsep{1.5pt}
% {@{}lcccccccc@{}}
\begin{tabular}{llccccccc}
\toprule
Dataset & Model & Baseline & Cutout & AA & PBA & FAA & Ours & Ours+AA\\
\hline
% \multicolumn{7}{c}{{\bf Reduced CIFAR-10}}\\
CIFAR-10 & Wide-ResNet-28-10 & 3.9 & 3.1 & 2.6 & 2.6 & 2.7 & 2.4 & {\bf 2.0} \\
\hline 
CIFAR-100 & Wide-ResNet-28-10 & 18.8 & 18.4 & 17.1 & 16.7 & 17.3 & 16.6 & {\bf 16.3} \\
\hline 
ImageNet & ResNet-50 & 23.7 & - & 22.4 & - & 22.4 & 22.5 & {\bf 22.0}\\
% \multirow{2}{*}{ImageNet} \\
\bottomrule
\end{tabular}
\end{center}
% \vspace{-2em}
% \vspace{-5mm}
\end{table}

\begin{figure}[t!]
\setlength{\abovecaptionskip}{-0.mm}
\setlength{\belowcaptionskip}{-0.mm}
\centering
\includegraphics[width=0.95\linewidth]{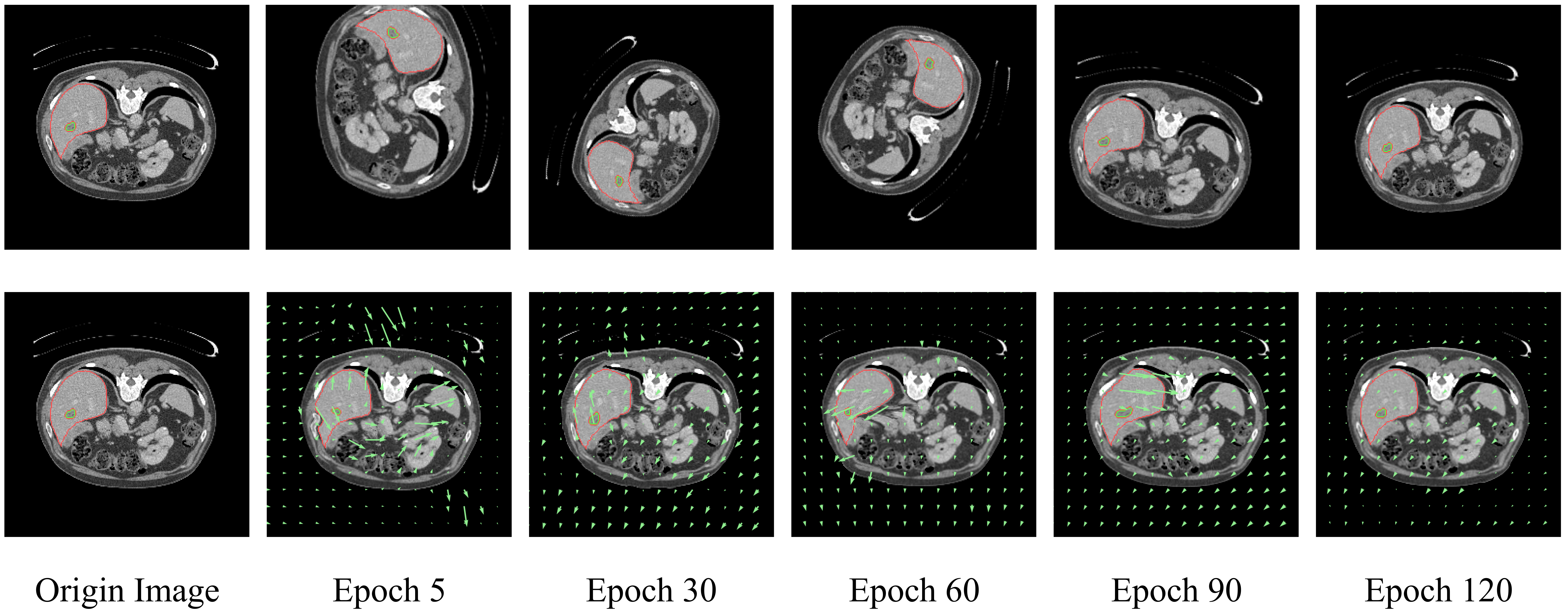}
\caption{Visualization of two augmentation modules: A-STN (\textbf{top}) and D-VAE (\textbf{bottom}). Red line is the contour of liver while green line is the contour of tumor. Our OnlineAugment can generate diverse augmented images. Moreover, it can also adapt to the target network. As the target network converges during training, the magnitude of the augmentation will also decrease. }
\label{fig:aug_vis}
\setlength{\belowdisplayskip}{3pt}
\end{figure}

\textbf{Comparisons with RandAugment on LiTS dataset.}
Our method has also achieved promising results on the semantic segmentation task, see in Table \ref{tab:comparison_lits}. The proposed method has comparable performance with RandAugment using global spatial transformation and has better results using local deformation. Moreover, OnlineAugment has a further performance boost by combining these two modules, while RandAugment does not benefit from the combination. Specifically, our onlineAugment achieves better results on the tumor images. It is because that OnlineAugment can adaptively generate harder samples for the target network, not just random ones. In Fig. \ref{fig:aug_vis}, augmented images of A-STN (top), and D-VAE (bottom) along with the training process are shown. One observation is that the predicted augmentation magnitude is relatively large at the beginning of training. As the training progress goes on, the network converges, and the predicted magnitude of augmentation gradually decreases.

%Compared with natural images, due to privacy and imaging difficulty, it is difficult to collect large amount of data for training neural networks in medical image analysis tasks. For medical image segmentation tasks, the annotation of data requires high expertise of doctors. Moreover, pixel-wise annotation is tedious, time-consuming and and labor-intensive. These factors combined together makes the cost of obtaining a large amount of medical segmentation data for training the network to be very high.  Therefore, effectively using a small amount of data to train an algorithm is an important issue for medical tasks. Data augmentation is one possible solution that can not only ease overfitting on small dataset, but also increase the generalization ability of the algorithm on unseen data.

\begin{table}[t!]
\begin{center}
\setlength{\abovecaptionskip}{-0.mm}
\setlength{\belowcaptionskip}{-0.mm}
\caption{Dice score coefficient comparison (\%) of our OnlineAugment with RandAugment on LiTS dataset. Our OnlineAugment has comparable performance with RandAugment in STN, and has much better results in local deformation and the combination of these two.}\label{tab:comparison_lits}
\setlength{\tabcolsep}{7mm}

\begin{tabular}{lccc}

\toprule
Method                    &Liver              &Tumor              & Average   \\
                       
\hline
no Augmentaion            &89.04             &44.73             &66.88  \\
\hline
RandAug STN               &\textbf{93.86}    &50.54             &72.20  \\
RandAug Deformation       &90.49             &46.93             &68.71  \\
RandAug Combine           &91.91             &51.11             &71.51   \\
\hline
Ours A-STN                &92.01             &52.26             &72.13   \\
Ours D-VAE                &90.18             &50.81             &70.49   \\
Ours Combine              &93.12             &\textbf{53.58}    &\textbf{73.35}\\

\bottomrule
\end{tabular}
\end{center}
% \vspace{-6mm}
\end{table}

% Visualization of augmented images and corresponding labels along different training epochs. First row is the results of STN module, and second row is the results of local deformation module.
\section{Conclusion}
In this paper, we have presented OnlineAugment - a new and powerful data augmentation technique.  Our method adapts online to the learner state throughout the entire training. We have designed three new augmentation networks that are capable of learning a wide variety of local and global geometric and photometric transformations, requiring less domain knowledge of the target task. Our OnlineAugment integrates both adversarial training and meta-learning for efficient training. We also design essential regularization techniques to guide adaptive online augmentation learning. Extensive experiments demonstrate the utility of the approach to a wide variety of tasks, matching (without requiring expensive offline augmentation policy search) the performance of the powerful AutoAugment policy, as well as improving upon the state-of-the-art in augmentation techniques when used jointly with AutoAugment. 

% In depth ablation study investigates the various aspects and components of our proposed approach.

% Our method is \textit{domain agnostic}, in a sense that it does not require an expensive offline augmentation search process to enter a new domain. In addition, o
% and generates the most effective augmentations it needs
% Our augmentation model architecture is based on
% augmentation learning network components

% We propose an efficient meta-learning framework for training OnlineAugment alongside the main learning task, 

% Future work directions include investigating the uses of OnlineAugment in a wide range of applications that could benefit from the ability to perform adaptive augmentation such as: test time augmentation, domain transfer, semi-supervised and unsupervised learning, as a defense against adversarial attacks and for measuring prediction uncertainty.
\section{Acknowledgment}
 This work has been partially supported by  NSF 1763523, 1747778, 1733843 and 1703883 Awards to Dimitris Metaxas and the Defense Advanced Research Projects Agency (DARPA) under Contract No. FA8750-19-C-1001 to Leonid Karlinsky and Rogerio Feris. %Any opinions, findings and conclusions or recommendations expressed in this material are those of the author(s) and do not necessarily reflect the views of DARPA.

% the Defense Advanced Research Projects Agency (DARPA).

\clearpage
\bibliographystyle{splncs04}
\bibliography{egbib}
\end{document}